\documentclass{ecai}
\usepackage{times}
\usepackage{graphicx}
\usepackage{latexsym}
\usepackage{amssymb}

\begin{document}

\title{Configuring Multiple Instances with Multi-Configuration}

\author{Alexander Felfernig\institute{Applied Software Engineering \& AI, Graz University of Technology, email: \{firstname.lastname\}@ist.tugraz.at} \and Andrei Popescu$^1$ \and Mathias Uta\institute{Siemens Energy AG, email: mathias.uta@siemens.com} \and  Viet-Man Le$^1$ \and \\ Seda Polat-Erdeniz$^1$  \and Martin Stettinger$^1$ \and Müslüm Atas$^1$ \and Thi Ngoc Trang Tran$^1$  }

\maketitle
\bibliographystyle{ecai}

\begin{abstract}
  Configuration is a successful application area of Artificial Intelligence. In the majority of the cases, configuration systems focus on configuring one solution (configuration) that satisfies the preferences of a single user or a group of users. In this paper, we introduce a new configuration approach -- multi-configuration -- that focuses on scenarios where the outcome of a configuration process is a set of configurations. Example applications thereof are the configuration of personalized exams for individual students, the configuration of project teams, reviewer-to-paper assignment, and hotel room assignments including individualized city trips for tourist groups. For multi-configuration scenarios, we exemplify a constraint satisfaction problem representation in the context of configuring exams. The paper is concluded with a discussion of open issues for future work.
\end{abstract}

\section{Introduction}
Configuration is a special case of design activity where a product is composed of a \emph{selection of predefined components} that satisfies a set of constraints \cite{Felfernigetal2014,sabin98-is,Stumptner1997}. Typical configurator applications are based on the assumption that a product (or service) is configured for a \emph{single user}. Related example applications can be found in product domains such as telecommunications \cite{fleischanderl1998}, automotive \cite{Landahl2014}, and software systems \cite{Sincero2008}. In contrast to the configuration for an individual user, \emph{group-based configuration} \cite{FelfernigAtasTranStettinger2016} is based on the idea of configuring a product or service for a group of users, i.e., the resulting configuration must take into account as much as possible the individual preferences of group members. An example of a group-based  scenario is the configuration and also recommendation of  software release plans where software requirements have to be arranged in such a way that the preferences of individual stakeholders are taken into account as much as possible \cite{FelfernigPrioritization2021,Felfernigetal2018}.

In this paper, we focus on a  scenario where \emph{multiple instances of the same product type} are configured for one user or a group of users. In this context, a set of constraints defines restrictions regarding the possible combinations of individual instances, i.e., \emph{the configured instances are not completely independent}. An application scenario for multi-configuration is the configuration of user-individual exams which have to take into account a set of constraints, for example, \emph{the share of complex questions per student must be below 20\%}. Following this  representation, we define \emph{multi-configuration} as \emph{a specific type of configuration task where multiple instances of the same type are configured for a single user or a group of users}. 

There are various example scenarios which can profit from multi-configuration. (1) When assigning presentation topics to students, the goal could be to identify three different starting paper references per student where, for example, \emph{each reference should be assigned to not more than two students}. (2) When planning room assignments for larger user groups (e.g., for a soccer training camp), a configurator could collect the preferences of individual group members and then generate a room assignment that takes into account as much as possible the preferences of the individual group members. (3) When configuring project teams, a configurator could take care of the optimal configuration of individual teams while keeping in mind the aspect of fairness, i.e., none of the team configurations should be "suboptimal". (4) When designing an apartment/house complex, neighborhood buildings must take into account rules such as \emph{the roof type of all buildings should be the same} or \emph{neighborhood buildings must not reduce the amount of direct sunlight below a specific treshold}. Compared to the former ones, the latter scenario can be regarded as a kind of industrial multi-configuration setting. Finally, the automated generation of test cases \cite{Gotlieb1998}, for example, in the context of regression testing, can be regarded as a multi-configuration task.

\section{Multi-Configuration}

In single user and group-based configuration scenarios, one configuration is determined that supports the requirements of the user (or a group of users). In multi-configuration scenarios, a set of configurations is determined for one or a group of users. The following discussions are based on a constraint-based configuration knowledge representation \cite{Tsang1993}. The following definition of a configuration task (and solution) allows  to represent collections of configurations  (see Definition 1 and Definition 2). A \emph{multi-configuration task} can be defined as a constraint satisfaction problem (CSP) \cite{Tsang1993}. 

\emph{Definition 1}. A \emph{Multi-Configuration Task} can be defined by a tuple $(V,D,REQ,C)$ where $V=\bigcup\{v_{ij}\}$ is a set of finite domain variables ($v_{ij}$ represents variable $j$ of configuration instance $i$), $D=\bigcup \{dom(v_{ij})\}$ is a set of corresponding domain definitions, $REQ=\bigcup \{v_{ij}=val_{ij}\}$ represents a set of user requirements, and $C=\{c_1, c_2, .., c_m\}$ is a set of constraints that restrict the way in which individual variable values can be combined with each other.

For simplicity, we assume that user requirements in $REQ$ are represented by simple variable value assignments ($v_{ij}=val_{ij}$). However, if needed, this assumption can be replaced by the more general case of allowing users to specify more complex constraints where each of those constraints is regarded as an individual requirement. Furthermore, we decided not to differentiate between different constraint types, for example, constraints referring to individual configurations and constraints referring to a set of configurations. On the basis of Definition 1, we can introduce the definition of a \emph{multi-configuration} (see Definition 2).

\emph{Definition 2}. A multi-configuration for a  multi-configuration task  $(V,D,REQ,C)$ is a set of variable value assignments $CONF=\bigcup \{v_{ij}=val_{ij}\}$ s.t.  $CONF \cup C \cup REQ$ is consistent. 

We now introduce an example \emph{multi-exam configuration} scenario.\footnote{For exam configuration, we replace the  term \emph{user} with \emph{examinee}/\emph{instructor}.}

\section{Example Multi-Configuration Scenario}

 In multi-exam configuration, the overall goal is to determine an exam instance for each examinee where a couple of constraints have to be taken into account. The underlying idea is to provide variability models for the purpose of being able to generate examinee-individual exams and thus save time in exam preparation and also avoid cheating due to exam diversity. Furthermore, this approach provides the possibility of a deeper integration of examinees into exam-related decision processes, i.e., students can be allowed to some extend to articulate their preferences regarding an exam. Following Definition 1, we now show how to represent the task of \emph{multi-exam configuration}.

\begin{itemize}
    \item $V=\{q_{11}..q_{kl}, q_{11}.type..q_{kl}.type, q_{11}.level.. q_{kl}.level\}$ where $q_{ij}$ is question $j$ posed to examinee $i$, $q_{ij}.type$ denotes the question type (category), $q_{ij}.level$ is the question complexity, $k$ the number of examinees, and $l$ the number of questions / examinee.
    \item $D=\{dom(q_{11})..dom(q_{kl}),dom(q_{11}.type)..dom(q_{kl}.type),$ $dom(q_{11}.level)..dom(q_{kl}.level)\}$, where $dom(q_{ij})=\{1..p\}$, $dom(q_{ij}.type)=\{1..\overline{q}\}$, and $dom(q_{ij}.level)=\{1..r\}$ ($p$ = number of questions per examinee, $\overline{q}$ = number of question categories, and $r$ = number of question complexity levels).
    \item $REQ=\{r_1 .. r_u\}$ where $r_\alpha$ is a requirement identifier and  $u$  the number of  requirements (defined by examinees and instructors).
    \item $C=\{c_1 .. c_v\}$ where $v$ is the number of constraints.
\end{itemize}


In the following, we introduce a couple of \emph{example constraints} that could be defined in the context of an exam configuration task.

\emph{Requirements}. $REQ$ includes a set of examinee- (and instructor-) individual constraints -- in the context of our example, $REQ$ is a set of examinee-specific constraints that have to be taken into account. The motivation behind this is that we want to make exam generation more flexible in terms of being able to include the preferences of examinees as well as the preferences of instructors.

First, we assume that examinee $a$ prefers to have included $\leq$ 30\% of questions related to the categories $\{\alpha,\beta\}$ (see Formula \ref{req1}). This way, we want to bring more flexibility into exam configuration, however, students will not always get what they want, i.e., their preferences have to be consistent with constraints defined by instructors. In many scenarios, students will not be allowed to define such constraints, i.e., only instructor requirements are relevant.

\begin{equation} \label{req1}
    r_1: \frac{|\{q_{aj} \in V: q_{aj}.type \in \{\alpha,\beta\}\}|}{|\{q_{aj} \in V\}|} \leq 0.3
\end{equation}

Furthermore, we assume that examinee $a$ prefers not to have included questions related to  question category $\gamma$ (see Formula \ref{req2}).

\begin{equation}  \label{req2}
    r_2:  |\{q_{aj} \in V: q_{aj}.type = \gamma\}|=0
\end{equation}

\emph{Question-level Constraints}. These constraints define properties related to the inclusion of specific questions. For example, we  assume that a specific question $v$ must not be included in more than $x$ exams (see Formula \ref{constraint1}), i.e., $x=0$ would exclude $v$ from any exam.

\begin{equation} \label{constraint1}
    c_1: |\{q_{ij} \in V: q_{ij}=v\}| \leq x
\end{equation}

Furthermore, we assume that each exam configuration includes question $u$ or question $v$ (see Formula \ref{constraint2}).

\begin{equation}  \label{constraint2}
    c_2: \bigwedge_{i=1}^{k(\#examinees)} \bigvee_{j=1}^{l(\#questions)} (q_{ij}=u \lor q_{ij}=v)
\end{equation}

We also assume that the minimum question complexity  level is $\pi$.

\begin{equation}
    c_3: \forall q_{ij} \in V: q_{ij}.level \geq \pi
\end{equation}

\emph{Global Constraints}. We  require that no question of type $\delta$ must be included in the exam (see Formula \ref{constraint3}). A simple reason for formulating such constraints could be that the related topic has not been discussed in detail within the scope of the course.

\begin{equation} \label{constraint3}
    c_4: |\{q_{ij} \in V: q_{ij}.type = \delta\}|=0
\end{equation}

We want to assure that at least three questions of type $\gamma$ must be included in each exam $u$ (see Formula \ref{constraint4}).

\begin{equation} \label{constraint4}
    c_5: \forall u \in \{1..k\}: |\{q_{uj} \in V: q_{uj}.type = \gamma\}| \geq 3
\end{equation}

Furthermore, the overall estimated duration of each exam must be $\Delta$ minutes (constraint formulated for examinee $a$) (see Formula \ref{constraint5}).

\begin{equation} \label{constraint5}
    c_6: \Sigma_{j=1}^{l(\#questions)} (q_{aj}.duration) = \Delta
\end{equation}

We also want to assure that the share of complex questions ($level = \phi$) per examinee must be between $16\%$ and $18\%$ (example formulated for examinee $a$) (see Formula \ref{constraint6}).

\begin{equation} \label{constraint6}
    c_7: 0.16 \leq \frac{|\{q_{aj} \in V: q_{aj}.level = \phi\}|}{|\{q_{aj} \in V\}|} \leq 0.18
\end{equation}

\emph{Further Constraints}. In addition to the mentioned examples, there are many further relevant constraints, for example, \emph{each examinee should be asked a specific question only once} and \emph{the question overlap between each pair of students should be more than 90\%}.

\section{Conclusions and Future Work}
We have introduced the notion of \emph{multi-configuration} which is a specific approach focusing on scenarios where collections of configurations are designed for user groups. In order to better understand the discussed concepts, we have introduced a working example from the domain of \emph{exam configuration}. Future work will include the analysis of the applicability of the presented concepts in the exam configuration domain (e.g., we will identify a complete set of typically relevant domain constraints) as well as in further multi-configuration scenarios. Furthermore, we will analyze new user interfaces and interaction requirements triggered by the application of multi-configuration concepts. The knowledge representation concepts discussed within the context of our exam configuration scenario are currently integrated into the \textsc{KnowledgeCheckr}  elearning environment (www.knowledgecheckr.com) \cite{StettingerCheckR2020}. Our major motivation is to increase the flexibility of exam generation  but also to counteract cheating in online exams through an increased exam variability. In cases where individual user requirements induce an inconsistency with the exam model constraints, we propose the application of model-based diagnosis concepts \cite{FelfernigetalFastDiag2012,LeetalICSE2021,Reiter1987} which can help to determine minimal conflict resolutions that also take into account aspects such as fairness and representativeness of the remaining questions.

\ack The presented work has been conducted in the \textsc{ParXCel} project funded by the Austrian Research Promotion Agency (880657).

\bibliography{bibliography}

\begin{thebibliography}{10}

\bibitem{FelfernigPrioritization2021}
A.~Felfernig, `{AI Techniques for Software Requirements Prioritization}', in
  {\em Artificial Intelligence Methods for Software Engineering}, eds.,
  M.~Kalech, R:~Abreu, and M.~Last,  29--47, World Scientific, (2021).

\bibitem{FelfernigAtasTranStettinger2016}
A.~Felfernig, M.~Atas, T.~Tran, and M.~Stettinger, `Towards group-based
  configuration', in {\em ConfWS'16}, pp. 69--72, Toulouse, France, (2016).

\bibitem{Felfernigetal2018}
A.~Felfernig, L.~Boratto, M.~Stettinger, and M.~Tkalcic, {\em Group Recommender
  Systems}, Springer, 2018.

\bibitem{Felfernigetal2014}
A.~Felfernig, L.~Hotz, C.~Bagley, and J.~Tiihonen, {\em Knowledge-based
  Configuration - From Research to Business Cases}, Elsevier, 2014.

\bibitem{FelfernigetalFastDiag2012}
A.~Felfernig, M.~Schubert, and C.~Zehentner, `{An Efficient Diagnosis Algorithm
  for Inconsistent Constraint Sets}', {\em AI for Engineering Design, Analysis,
  and Manufacturing}, {\bf 26}(1),  53--62, (2012).

\bibitem{fleischanderl1998}
G.~Fleischanderl, G.~Friedrich, A.~Haselboeck, H.~Schreiner, and M.~Stumptner,
  `Configuring large systems using generative constraint satisfaction', {\em
  IEEE Intelligent Systems}, {\bf 13}(4),  59--68, (1998).

\bibitem{Gotlieb1998}
A.~Gotlieb, B.~Botella, and M.~Rueher, `{Automatic Test Data Generation Using
  Constraint Solving Techniques}', {\em ACM SIGSOFT Software Engineering
  Notes}, {\bf 23}(2),  53--62, (1998).

\bibitem{Landahl2014}
J.~Landahl, D.~Bergsjö, and H.~Johannesson, `{Future Alternatives for
  Automotive Configuration Management}', {\em Procedia Computer Science}, {\bf
  28},  103--110, (2014).

\bibitem{LeetalICSE2021}
V.M. Le, A.~Felfernig, M.~Uta, D.~Benavides, J.~Galindo, and T.N.T. Tran,
  `{\textsc{DirectDebug}: Automated Testing and Debugging of Feature Models}',
  in {\em 43rd Intl. Conference on Software Engineering (ICSE'21)}, pp. 81--85,
  Virtual, (2021). IEEE.

\bibitem{Reiter1987}
R.~Reiter, `A theory of diagnosis from first principles', {\em Artificial
  Intelligence}, {\bf 32}(1),  57--95, (1987).

\bibitem{sabin98-is}
D.~Sabin and R.~Weigel, `{Product Configuration Frameworks - A Survey}', {\em
  IEEE Intelligent Systems}, {\bf 13}(4),  42--49, (1998).

\bibitem{Sincero2008}
J.~Sincero and W.~Schröder-Preikschat, `{The Linux Kernel Configurator as a
  Feature Modeling Tool}', in {\em Workshop on Analyses of Software Product
  Lines}, pp. 257--260, Limerick, Ireland, (2008). ASPL.

\bibitem{StettingerCheckR2020}
M.~Stettinger, T.N.T. Tran, I.~Pribik, G.~Leitner, A.~Felfernig, R.~Samer,
  M.~Atas, and M.~Wundara, `{KNOWLEDGECHECKR: Intelligent Techniques for
  Counteracting Forgetting}', in {\em {24th European Conference on Artificial
  Intelligence (ECAI 2020)}}, eds., G.~DeGiacomo, A.~Catala, B.~Dilkina,
  M.~Milano, S.~Barro, A.~Bugarin, and J.~Lang, volume 325 of {\em Frontiers in
  Artificial Intelligence and Applications}, pp. 3034--3039, Santiago de
  Compostela, Spain, (2020). IOS Press.

\bibitem{Stumptner1997}
M.~Stumptner, `An overview of knowledge-based configuration', {\em AI
  Communications}, {\bf 10}(2),  111--125, (1997).

\bibitem{Tsang1993}
E.~Tsang, {\em Foundations of Constraint Satisfaction}, Academic Press, 1993.

\end{thebibliography}
\end{document}